# Architectural Flaw Detection in Civil Engineering Using GPT-4

Saket Kumar
*The Mathworks Inc*
saketk@mathworks.com

Abul Ehtesham
*The Davey Tree Expert Company*
abul.ehtesham@davey.com

Aditi Singh
*Department of Computer Science*
*Cleveland State University*
a.singh22@csuohio.edu

Tala Talaei Khoei
*Khoury College of Computer Science*
*Roux Institute at Northeastern University*
t.talaeikhoei@northeastern.edu

*Abstract*—The application of artificial intelligence (AI) in civil engineering presents a transformative approach to enhancing design quality and safety. This paper investigates the potential of the advanced LLM GPT4 Turbo vision model in detecting architectural flaws during the design phase, with a specific focus on identifying missing doors and windows. The study evaluates the model's performance through metrics such as precision, recall, and F1 score, demonstrating AI's effectiveness in accurately detecting flaws compared to human-verified data. Additionally, the research explores AI's broader capabilities, including identifying load-bearing issues, material weaknesses, and ensuring compliance with building codes. The findings highlight how AI can significantly improve design accuracy, reduce costly revisions, and support sustainable practices, ultimately revolutionizing the civil engineering field by ensuring safer, more efficient, and aesthetically optimized structures.

## I. Introduction

Civil engineering is at the core of infrastructure development, encompassing the design, construction, and maintenance of buildings, bridges, roads, and other large structures. As projects grow in complexity and scale, the potential for design and structural flaws increases, potentially leading to catastrophic failures, cost overruns, and extended timelines. Traditional methods of detecting these flaws often rely on manual inspection, human expertise, and rudimentary software tools, which may miss subtle yet critical issues.

The advent of Artificial Intelligence (AI) presents an opportunity to significantly enhance the accuracy and efficiency of flaw detection in civil engineering. By leveraging advanced Large Language models [1] like GPT-4 Multimodal [2], LLaMA and Gemini Pro vision model, engineers can harness vast computational power to analyze complex data sets, identify patterns, and predict potential problems before they manifest. This paper explores how AI can be applied to detect a wide range of architectural and structural flaws, ultimately improving the safety, sustainability, and overall quality of civil engineering projects.

## II. AI in Detecting Load-Bearing Issues and Material Weaknesses

Load-bearing elements are critical to the stability and safety of any structure. AI can analyze design models, including Building Information Modeling (BIM) data, to assess whether the load distribution is appropriate for the intended use. For example, AI can simulate various load scenarios, such as static and dynamic loads, to identify areas where structural components may fail under stress. This early detection allows engineers to modify designs before construction begins, reducing the risk of failure.

Material weaknesses, whether due to poor-quality materials or improper use, are another significant concern in civil engineering. AI can assist by analyzing data from material tests, such as tensile strength, compressive strength, and elasticity, to ensure that the materials selected will perform as required under different conditions. Additionally, AI can monitor materials during construction to detect deviations from specified standards, ensuring that the integrity of the structure is maintained.

## III. AI in Ensuring Compliance with Building Codes and Standards

Building codes and standards are designed to ensure the safety and functionality of structures, but compliance can be complex and time-consuming. AI can streamline this process by automatically cross-referencing design models with relevant codes and standards. For example, AI can check whether the design meets fire safety regulations, such as the placement of fire exits and the use of fire-resistant materials. Similarly, AI can ensure that the design complies with accessibility standards, identifying potential issues such as inadequate ramp slopes or insufficient space for wheelchair access. By automating compliance checks using Agentic workflows [3], AI reduces the likelihood of human error and ensures that all aspects of the design meet the required standards. This not only enhances safety but also reduces the time and cost associated with manual reviews and revisions.

## IV. AI in Identifying Ergonomic and Accessibility Problems

Ergonomics and accessibility are increasingly important considerations in civil engineering, particularly in public and commercial buildings. AI can analyze design models to identify potential issues that may affect the comfort, safety, or usability of a space. For example, AI can assess the layout of a building to ensure that it is intuitive and easy to navigate or evaluate whether the design provides adequate space for people with disabilities.

In addition, AI can simulate how different user groups will interact with space, allowing engineers to optimize the design for various needs. This can include analyzing the placement of furniture, lighting, and ventilation to create a more comfortable and functional environment.

## V. AI in Assessing Fire Safety Hazards and Seismic Vulnerabilities

Fire safety and seismic resilience are critical considerations in civil engineering, particularly in regions prone to natural disasters. AI can enhance these aspects by analyzing design models to identify potential fire hazards, such as insufficient fire exits, poorly placed fire alarms, or the use of flammable materials. AI can also simulate the spread of fire through a building, allowing engineers to optimize the design for fire safety.

Similarly, AI can assess a structure's seismic resilience by simulating how it would behave during an earthquake. This can include analyzing the design's ability to absorb and dissipate seismic energy, identifying weak points in the structure, and suggesting design modifications to enhance resilience.

## VI. AI in Evaluating Sustainability and Water Management

Sustainability is a growing concern in civil engineering, with increasing emphasis on reducing the environmental impact of construction projects. AI can assist by evaluating the sustainability of a design, including its energy efficiency, carbon footprint, and water management strategies [4]. For example, AI can analyze the design's ability to capture and reuse rainwater, reduce energy consumption, or use sustainable materials.

Water management is another critical aspect of sustainability, particularly in regions prone to flooding or drought. AI can model the impact of different water management strategies, such as the use of permeable surfaces or the creation of green spaces, on the overall sustainability of a project.

## VII. AI in Analyzing Construction Feasibility and Resource Optimization

Construction feasibility is a key concern in civil engineering, particularly for large or complex projects. AI can analyze the design and construction plans to assess whether they are feasible given the available resources, such as labor, materials, and equipment. This can include evaluating the timeline, budget, and logistics of the project to identify potential challenges and suggest alternative approaches.

Table I provides a summary of recent civil engineering failures, highlighting the causes behind these catastrophic events, such as long-term degradation, structural deficiencies, and poor maintenance. These examples underscore the importance of early detection and assessment of structural vulnerabilities, which AI systems can address effectively.

AI can also optimize resource use by analyzing the design and construction process to identify areas where materials, labor, or time can be saved. This can include optimizing the layout of a construction site, reducing waste, or identifying more efficient construction methods.

## VIII. AI in Predicting Maintenance Needs and Continuous Structural Health Monitoring

Predictive maintenance is another area where AI can make a significant impact. By analyzing data from sensors embedded in a structure, AI can predict when maintenance will be needed, reducing the likelihood of unexpected failures. This can include monitoring for signs of wear and tear, corrosion, or other issues that could affect the structure's integrity.

Continuous structural health monitoring [5] is another application of AI that can enhance the safety and longevity of a structure. By continuously analyzing data from sensors, AI can detect subtle changes in the structure's condition, such as shifts in load distribution or the development of cracks. This allows engineers to address issues before they become serious, reducing the risk of failure and extending the structure's lifespan.

## IX. System Architecture

Fig. 1 illustrates the system architecture for AI-driven flaw detection, consisting of four steps: (1) Images are accessed and converted to Base64 for processing, (2) The GPT-4 Turbo model analyzes these images to detect flaws, (3) Results are recorded, indicating whether flaws related to doors or windows were detected, and (4) Further analysis generates reports, including confusion matrices and statistical data for research. Fig. 2 depicts the prompt given to the AI model to for steps 2 and 3.

Table I. Recent Civil Engineering Failures and their Causes

| Failure Event | Year | Failure | Causes | References |
|---|---|---|---|---|
| Surfside Condominium Collapse | 2021 | Structural Collapse | Long-term degradation, poor maintenance, and ignored warnings | [6] |
| Genoa Morandi Bridge Collapse | 2018 | Structural Failure | Corrosion and structural deficiencies | [7] |
| Florida International University Pedestrian Bridge Collapse | 2018 | Structural Failure | Design flaws and inadequate load testing | [8] |
| Pittsburgh Fern Hollow Bridge Collapse | 2022 | Structural Collapse | Poor maintenance and structural deterioration | [9] |
| Gwangju Building Collapse | 2021 | Construction Failure | Unsafe demolition practices | [10] |
| Mexico City Metro Overpass Collapse | 2021 | Structural Failure | Structural fatigue, poor maintenance, and design flaws | [11] |
| Hard Rock Hotel Collapse (New Orleans) | 2019 | Construction Failure | Structural design flaws and improper construction techniques | [12] |

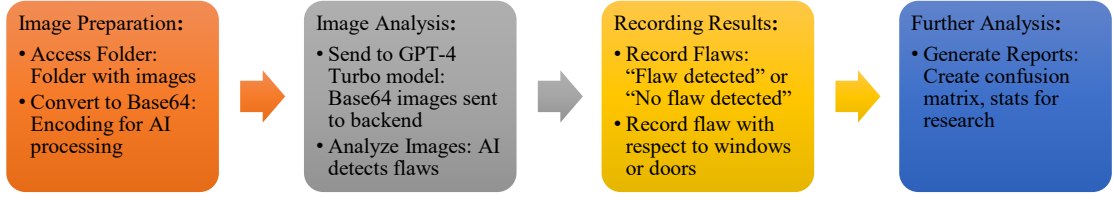

Fig. 1. Architecture Overview

Please identify if there are missing doors, windows, or walls in the floor plan image. Respond only with a JSON object containing the following keys:

"door": 1 if door is missing, otherwise 0,

"window": 1 if window is missing, otherwise 0,

"image_flaw_detected": 1 if any flaw is detected, otherwise 0.

Fig. 2. Prompt message to AI assistant

## X. RESULTS AND DISCUSSION

### A. Experiment

This study involved evaluating ChatGPT's vision model (ChatGPT4 Turbo) on a dataset of 100 floor plans to assess its ability to detect architectural flaws, specifically missing door and windows. The dataset used for the experiments in this study can be accessed at https://github.com/asinghcsu/GPTCivilArchFlaw. The models prediction were compared against human-verified ground truth data and results were analyzed using statistical metrics such as confusion matrices, recall, precision, F1-score and ROC curves.

- *Recall,* also known as Sensitivity or True Positive Rate measures the model's ability to correctly identify all positive instances. It is defined as the ratio of true positive predictions to the total actual positives and can be computed using Equation 1.

$$Recall = \frac{TP}{(TP+FN)} \quad (1)$$

- *Precision* quantifies the accuracy of the positive predictions made by the model. It is defined as the ratio of true positive predictions to the total predicted positives, as shown in Equation 2.

$$Precision = \frac{TP}{(TP+FP)} \quad (2)$$

- *F1-score* is the harmonic mean of precision and recall, providing a balanced measure between precision and recall, as shown in using 3.

$$F1-score = \frac{2*Precision*Recall}{(Precision+Recall)} = \frac{TP}{TP+\frac{1}{2}(FP+FN)} \quad (3)$$

- *AUC* refers to the area beneath the Receiver Operating Characteristic (ROC) curve, which plots the True Positive Rate against the False Positive Rate at various threshold settings. The AUC value can be calculated using the integral in Equation 4.

$$AUC = \int \frac{TP}{(TP+FN)} d\left(\frac{FP}{(FP+TN)}\right) \quad (4)$$

### B. Illustration of a Floor Plan

As part of this study, ChatGPT's vision model was tested on 100 floor plans to detect flaws such as missing doors and windows. The model's predictions were compared with human-confirmed results. Fig. 3 displays a floor plan of a small apartment, which includes a living/dining area, kitchen, bedroom, two toilets, a terrace, and a flower bed, along with the respective room dimensions. Fig. 4 presents another floor plan that exhibits flaws missing door and missing window in bedroom.

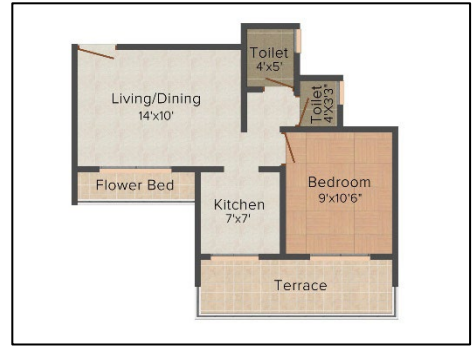

Fig. 3. Example floor plan

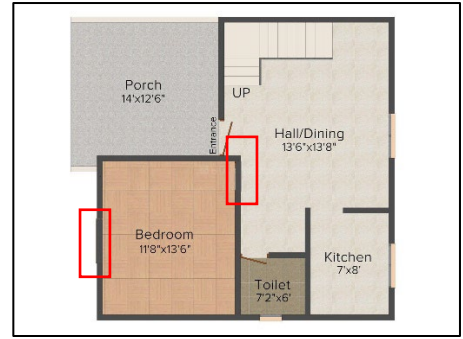

Fig. 4. Example floor plan with flaw

## C. Results

The experiment evaluated the model's performance in detecting flaws in the dataset of 100 floor plans. Various performance metrics, such as precision, recall, and F1-score, were calculated to assess the model's effectiveness. The confusion matrices and ROC curves ( Fig 5C, 6C, 7C) provided further insights into the model's capabilities.

- *Confusion Matrix (Flaw Detection):* The confusion matrix for flaw detection (Fig 5A) shows how well the model predicted flaws compared to the actual flaws confirmed by a human. The matrix reveals the counts of true positives, true negatives, false positives, and false negatives.

- *Model Performance Metrics (Flaw Detection):* The bar chart in Fig 5B illustrates the model's precision (0.65), recall (0.57), and F1-Score (0.61) in detecting flaws. These metrics suggest the model is relatively balanced, with strong precision and recall, indicating effective detection of flaws with few false positives and false negatives.

- *Confusion Matrix (Door Missing Detection):* Fig 6A shows the model's performance in predicting missing doors. The counts in the matrix provide a clear view of where the model succeeded and where it failed, highlighting any imbalances or misclassifications in this specific task.

- *Model Performance Metrics (Door Missing Detection):* The metrics shown in Fig 6B indicate the model's precision (0.33), recall (0.17), and F1-Score (0.23), in detecting missing doors. The lower recall suggests that the model missed some instances of missing doors, but it maintained a decent balance overall.

- *Confusion Matrix (Window Missing Detection):* The confusion matrix (Fig 7A) for window missing detection compares predicted versus actual cases of missing windows, showcasing the model's strengths and weaknesses in this task.

- *Model Performance Metrics (Window Missing Detection):* The bar chart (Fig 7B) indicates the model's precision (0.56), recall (0.68), and F1-Score (0.61) in detecting missing windows, with strong precision and a relatively balanced performance across all metrics.

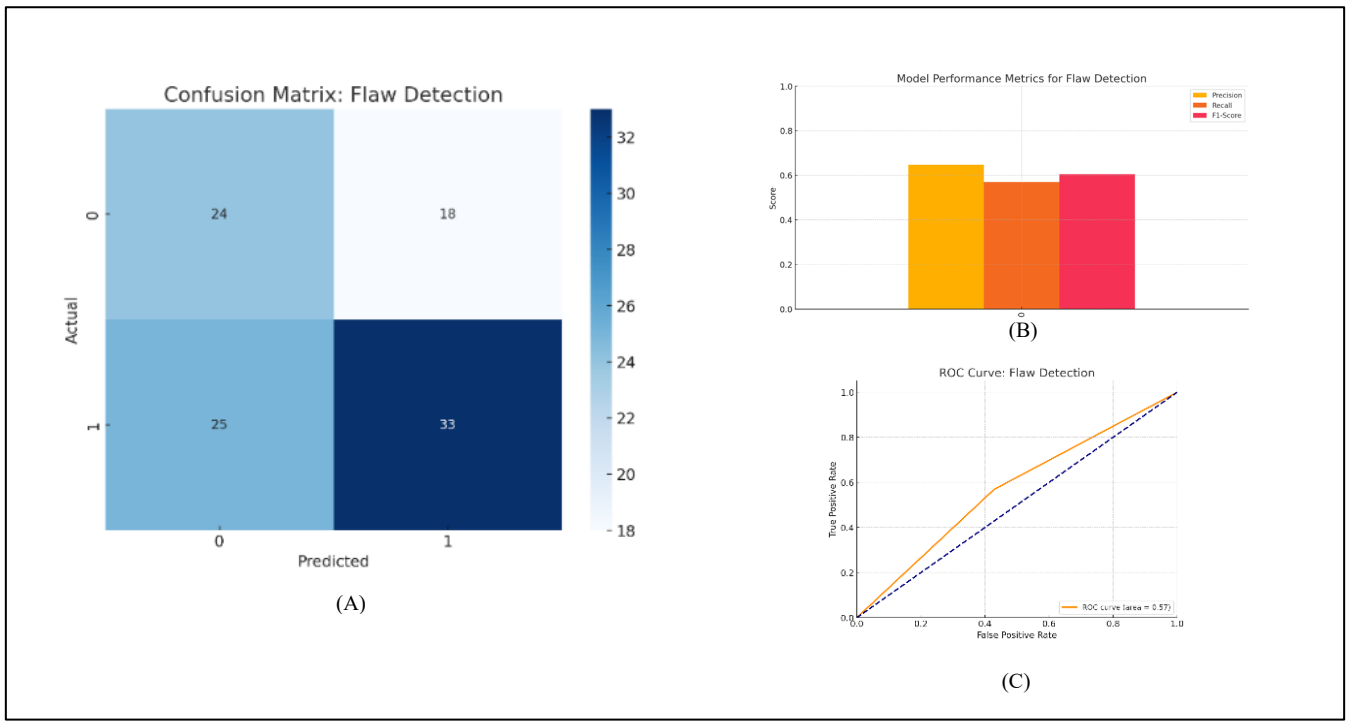

Fig. 5. Confusion Matrix, Model Performance Metrics and ROC Curve for Flaw Detection

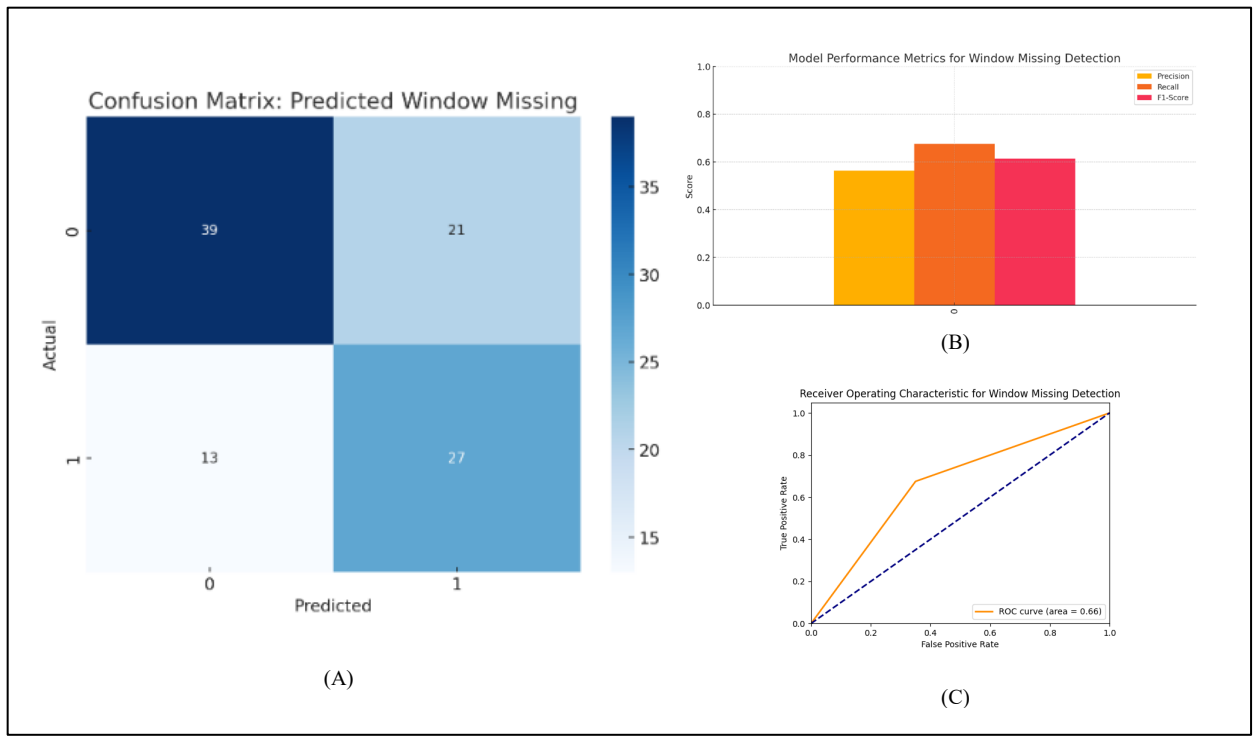

Fig. 6. Confusion Matrix, Model Performance Metrics and ROC Curve for Window Detection

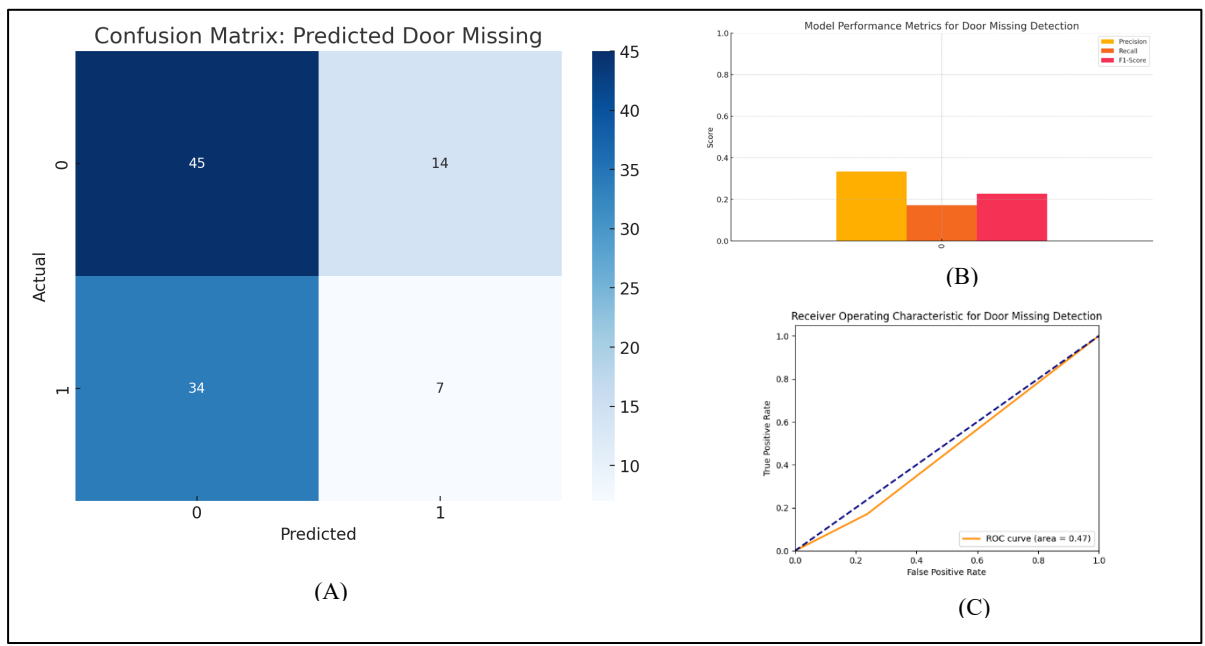

Fig. 7. Confusion Matrix, Model Performance Metrics and ROC Curve for Door Detection

## XI. CONCLUSION

The integration of Generative AI into civil engineering marks a significant advancement in the field, offering innovative tools and techniques for detecting architectural and structural flaws. By enhancing design accuracy, reducing costly revisions, and promoting sustainable practices, GenAI has the potential to revolutionize civil engineering projects from planning to execution and maintenance. The ability of GenAI to identify a wide range of issues—from load-bearing problems to sustainability concerns—ensures that structures are not only safe and efficient but also aesthetically pleasing and environmentally responsible.

As AI technology continues to evolve, its role in civil engineering is likely to expand, leading to even greater improvements in the quality and safety of our built environment. Future research could explore the integration of AI with other emerging technologies, such as augmented reality and real-time data analytics, to further enhance the design and monitoring processes. This continued innovation will be crucial in addressing the increasing complexity and demands of modern infrastructure development.

References

[1] A. Singh, "Exploring Language Models: A Comprehensive Survey and Analysis," *2023 International Conference on Research Methodologies in Knowledge Management, Artificial Intelligence and Telecommunication Engineering (RMKMATE)*, Chennai, India, 2023, pp. 1-4, doi: 10.1109/RMKMATE59243.2023.10369423.

[2] GPT-4 Turbo and GPT-4 Models," Available: https://platform.openai.com/docs/models/gpt-4-turbo-and-gpt-4. [Accessed: Aug. 30, 2024].

[3] A. Singh, A. Ehtesham, S. Kumar, and T. T. Khoei, "Enhancing AI Systems with Agentic Workflows Patterns in Large Language Model," *2024 IEEE World AI IoT Congress (AIIoT)*, Seattle, WA, USA, 2024, pp. 527-532, doi: 10.1109/AIIoT61789.2024.10578990.

[4] National Institute of Standards and Technology (NIST). (2022). *AI in Civil Engineering: Current Trends and Future Directions*.

[5] E. Figueiredo et al., "Machine Learning Algorithms for Damage Detection in SHM: Applications and Challenges," *Mechanical Systems and Signal Processing*, vol. 25, no. 7, pp. 2093-2122, 2011.

[6] National Institute of Standards and Technology, "Champlain Towers South Investigation Report," 2022. [Online]. Available: https://www.nist.gov/disaster-failure-studies/champlain-towers-south-collapse-ncst-investigation. [Accessed: 30-Aug-2024].

[7] Italian Ministry of Infrastructure and Transport, "Morandi Bridge Collapse Investigation Report," 2019.

[8] National Transportation Safety Board, "Collapse of Pedestrian Bridge Over SW 8th Street, Miami, Florida," National Transportation Safety Board Report, 2019. [Online]. Available: https://www.ntsb.gov/investigations/AccidentReports/Reports/HAR1902.pdf. [Accessed: 30-Aug-2024].

[9] National Transportation Safety Board, "Pittsburgh Fern Hollow Bridge Collapse Investigation," National Transportation Safety Board Report, 2022. [Online]. Available: https://www.ntsb.gov/investigations/AccidentReports/Reports/HIR2402.pdf. [Accessed: 30-Aug-2024].

[10] J. Lee, "Gwangju Building Collapse: A Case of Negligence," Yonhap News Agency, 2021.

[11] Reuters, "Mexico City Metro Overpass Collapse: Engineering Failures and Political Fallout," Reuters News Agency, 2021. [Online]. Available: https://www.reuters.com/world/americas/mexico-city-metro-accident-partly-due-lack-maintenance-third-audit-finds-2022-05-10/. [Accessed: 30-Aug-2024].

[12] Occupational Safety and Health Administration, "Investigation Report on the Hard Rock Hotel Collapse," 2020. [Online]. Available: https://www.osha.gov/news/newsreleases/region6/04032020. [Accessed: 30-Aug-2024].